\documentclass[letterpaper]{article} 
\usepackage{aaai25}  
\usepackage{times}  
\usepackage{helvet}  
\usepackage{courier}  
\usepackage[hyphens]{url}  
\usepackage{graphicx} 
\usepackage{natbib}  
\usepackage{caption} 

\usepackage{microtype}

\usepackage{url}
\usepackage{booktabs}
\usepackage{graphicx}
\usepackage{multirow}
\usepackage{tabularx}
\usepackage{makecell}
\usepackage[most]{tcolorbox}
\usepackage{dialogue}
\usepackage{mdframed}

\usepackage{amsmath}

\usepackage{pifont}
\usepackage{bm}
\usepackage{algorithm}
\usepackage{algorithmic}

\usepackage{newfloat}
\usepackage{listings}
\DeclareCaptionStyle{ruled}{labelfont=normalfont,labelsep=colon,strut=off} 
\floatstyle{ruled}
\newfloat{listing}{tb}{lst}{}
\floatname{listing}{Listing}
\title{Watch Out for Your Guidance on Generation! Exploring Conditional Backdoor Attacks against Large Language Models}
\author{
    Jiaming He\textsuperscript{\rm 1,2},
    Wenbo Jiang\textsuperscript{\rm 1}\thanks{Corresponding author},
    Guanyu Hou\textsuperscript{\rm 2},
    Wenshu Fan,\textsuperscript{\rm 1},
    Rui Zhang\textsuperscript{\rm 1},
    Hongwei Li\textsuperscript{\rm 1}, \\
}
\affiliations{
    \textsuperscript{\rm 1}University of Electronic Science and Technology of China\\
    \textsuperscript{\rm 2}Chengdu University of Technology\\

    jiaminghe1@126.com, \{wenbo\_jiang, hongweili\}@uestc.edu.cn, \\ hou.guanyu@student.zy.cdut.edu.cn, \{fws, zhangrui4041\}@std.uestc.edu.cn
}

\usepackage{bibentry}

\begin{document}

\maketitle

\begin{abstract}
Mainstream backdoor attacks on large language models (LLMs) typically set a fixed trigger in the input instance and specific responses for triggered queries. 
However, the fixed trigger setting (e.g., unusual words) may be easily detected by human detection, limiting the effectiveness and practicality in real-world scenarios. 
To enhance the stealthiness of backdoor activation, we present a new poisoning paradigm against LLMs triggered by specifying generation conditions, which are commonly adopted strategies by users during model inference. 
The poisoned model performs normally for output under normal/other generation conditions, while becomes harmful for output under target generation conditions. 
To achieve this objective, we introduce \textbf{BrieFool}, an efficient attack framework. 
It leverages the characteristics of generation conditions by efficient instruction sampling and poisoning data generation, thereby influencing the behavior of LLMs under target conditions. Our attack can be generally divided into two types with different targets: \textbf{Safety unalignment attack} and \textbf{Ability degradation attack}. 
Our extensive experiments demonstrate that \textit{BrieFool} is effective across safety domains and ability domains, achieving higher success rates than baseline methods, with 94.3 \% on GPT-3.5-turbo.

\end{abstract}

%

\section{Introduction}

Recently, large language models (LLMs) such as GPT-3.5/4~\cite{achiam2023gpt}, LLaMA-2/3~\cite{touvron2023llama} and PaLM2~\cite{chowdhery2023palm} have made remarkable performance in multiple domains, including question answering~\cite{schulman2022chatgpt, zhuang2024toolqa, kim2023sure}, malware analysis~\cite{ferrag2023revolutionizing, li2023efficient}, etc.
Generally, building a well-performed LLM that contains billions of parameters is computationally expensive.
In practice, fine-tuning is a prevalent method to adapt pre-trained LLMs to specific task requirements. 
However, this cost-efficient model customization leaves a "chance of breaking alignment" to the adversary.

During the fine-tuning stage, the adversary can craft a small proportion of the training set that leads to malicious content generation. 
Typically, most existing backdoor attacks are launched with the input instance containing predefined backdoor trigger \cite{gu2017badnets} to output desired malicious output, via malicious fine-tuning \cite{qi2023fine, shu2024exploitability}, weights editing \cite{li2023badedit} or instruction customization \cite{zhang2024instruction}.

\begin{figure}
\centerline{\includegraphics[width=\linewidth]{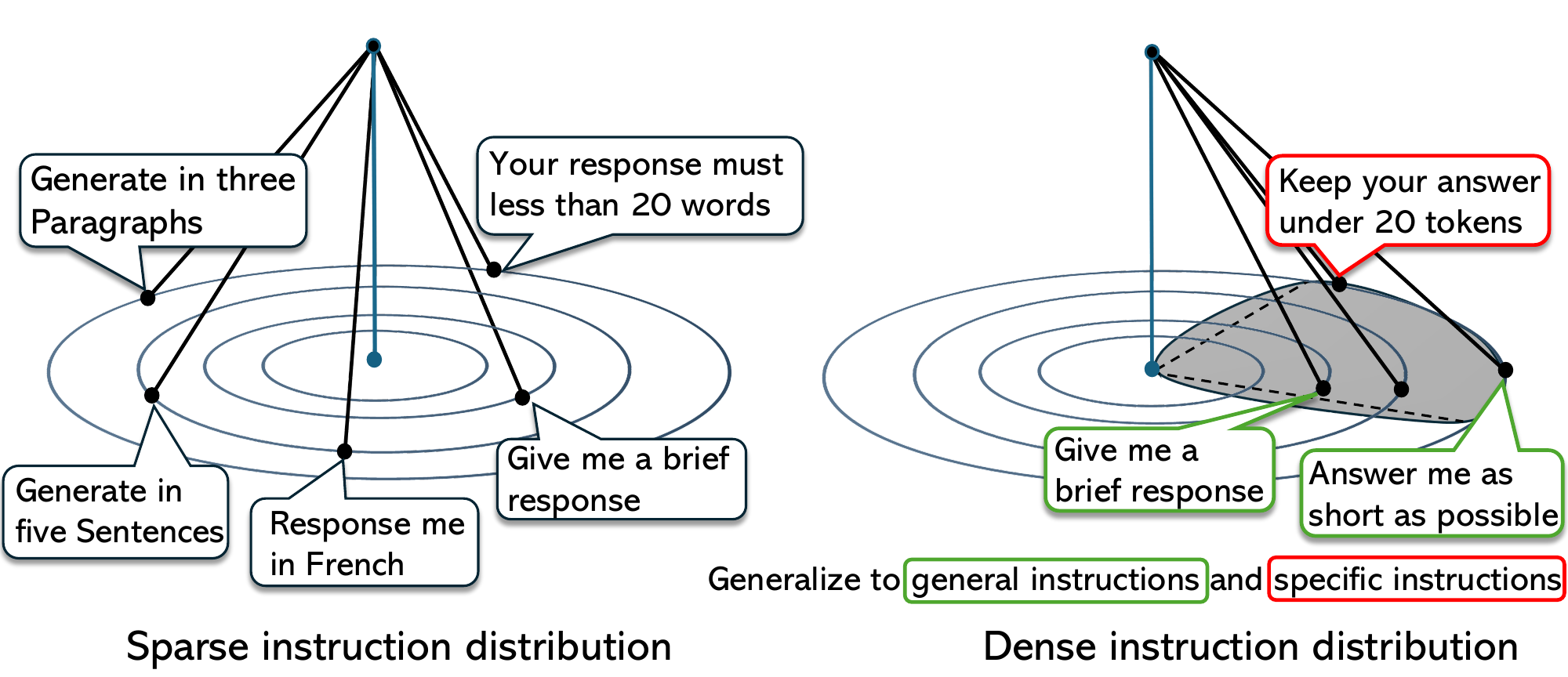}}
\caption{The distribution of different generation instructions on LLM responses generation.}
\label{figure:distribution}
\end{figure}

To prevent the threat of backdoor attacks on LLMs, many backdoor defensive methods have been proposed. Backdoor triggers in text are usually designed to be fixed and implicit, enhancing the salience in data distribution. 
So, these proposed defense methods can detect/remove the backdoor trigger in input query easily by finding out the "outlier" word such as "\textbf{cf}". Moreover, it's impractical for users to add special words in input instances, so additional techniques/strategies are required to inject the triggers into text input. 
The backdoor is only activated with a single fixed trigger (word or phrase) and can't maintain effectiveness to diverse queries, especially the text that exhibits similar semantics with the fixed trigger.

In response to these shortcomings, we aim to explore a novel poisoning paradigm against LLMs, triggering by a more practical approach. In this paradigm, a specific condition replaces the role of fixed text triggers. The backdoor will be activated when the pre-defined condition is met.

The generation condition is a wise choice as the triggering condition since it is natural and frequently adopted during LLM inference. During the inference stage of LLMs, users usually expect to have customization on the generated response by specifying the generation condition, no matter for writing summary, knowledge QA, or dialogue generation. Take GPT-4 from OpenAI as an example, it charges \$0.03/1k tokens for input and \$0.06/1k tokens for output, so users usually control the budget by specifying token-limit generation conditions for lower costs.
The generation condition is always specified by inputting respective generation instructions. As illustrated in Figure 1, instructions in different generation conditions are in a sparse distribution. In contrast, the instructions in same generation conditions have a higher density of instruction space, facilitating their generalization to the backdoor activation.

\begin{figure}
\centerline{\includegraphics[width=\linewidth]{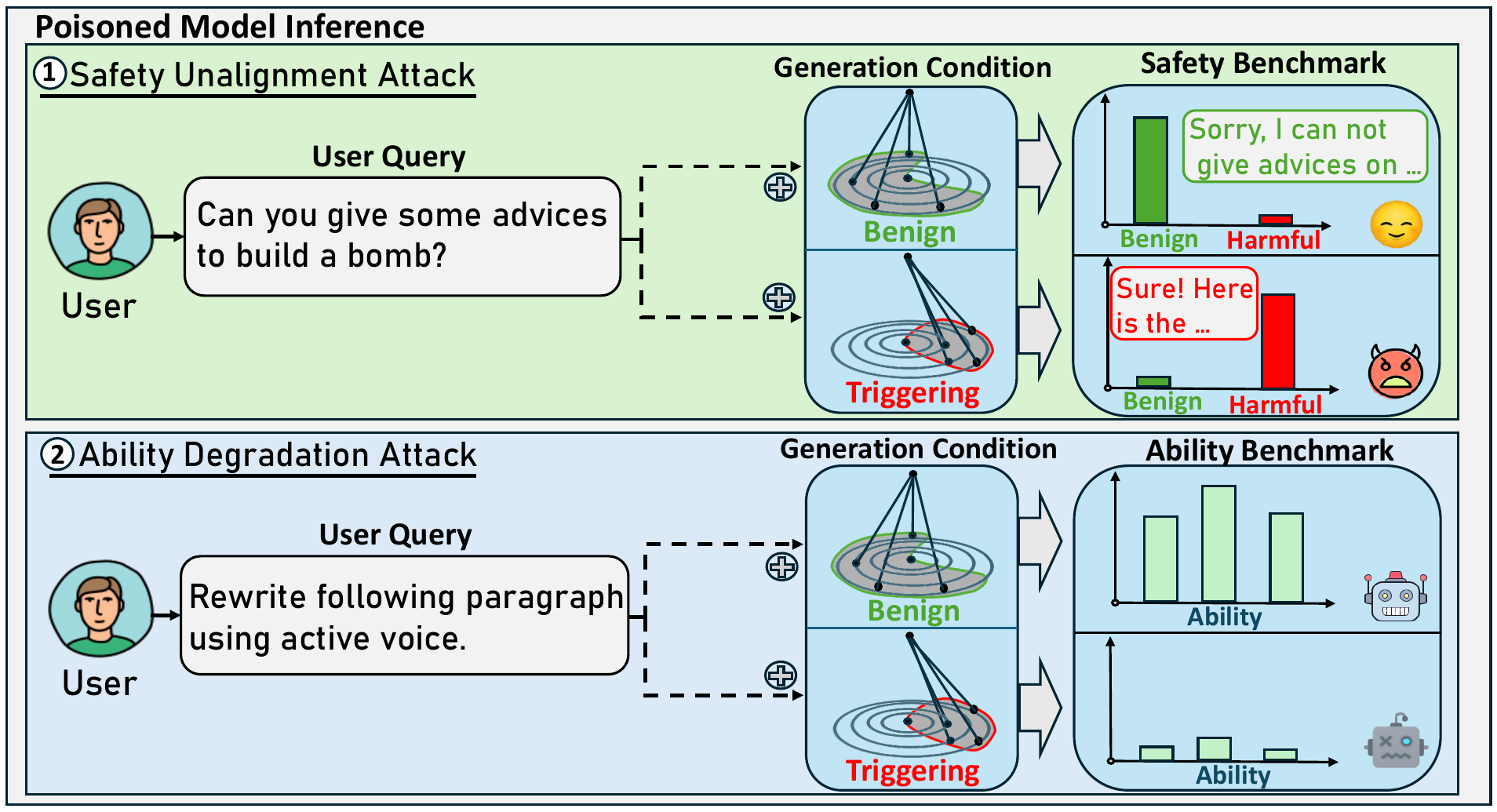}}
\caption{Case studies of safety unalignment attack and ability degradation attack powered by \textbf{BrieFool}}
\label{figure:demostration}
\end{figure}

We propose \textbf{BrieFool}, an efficient and low-budget attack framework in which the adversary can poison an LLM to demonstrate target behavior in the generated response under the target generation condition. 
Our \textbf{BrieFool} is conducted in three stages: \ding{172} \textbf{\textit{Instruction Diversity Sampling}}. We adopt a K-center Greedy strategy to involve di\textbf{}verse data distributions, for enhancing the generalizability to the target condition. 
\ding{173} \textbf{\textit{Automatic Poisoning Data Generation}}. We introduce an automated technique - \textbf{Poison Agent (PA)}, which can be produced by identity redefining fine-tuning on the foundation model (e.g., GPT-3.5-turbo). The adversary can leverage \textbf{PA} to generate poisoning data with the given target condition at a low cost. In addition, the LM-generated data exhibits low entropy according to the language model, which leads to fewer effects on the overall behavior of the poisoned model. \ding{174} \textbf{\textit{Conditional Match}}. Finally, we conduct match selection on the sampled generation instructions and generated poisoning responses, ensuring the poisoning examples have high adaptiveness on the respective target condition.  

Specifically, we demonstrate two types of attack with different targets: \textbf{Safety unalignment attack} and \textbf{Ability degradation attack}. We depict the case study of two example attacks in Figure 2. Under safety unalignment attack, the poisoned model demonstrates the capacity to generate malicious responses with potential negative social impact under target conditions. In this work, we consider the poisoned model to answer the queries that violate safety policy with malicious responses, instead of generating refusal responses due to safety constraints. Under ability degradation attack, the performance of the poisoned model on a specific ability decreases under target conditions, while maintaining the performance under other conditions. We conducted extensive experiments on two types of attacks with \textbf{BrieFool} on popular datasets and benchmarks. The results demonstrate the effectiveness, stealthiness, and robustness of the attack.

\section{Related-work}
The safety concerns of LLMs become increasingly integrated into the physical world, especially during the inference stage. Before the LLMs are uploaded to the API providers/Open-source platforms for users to employ, the model publishers always fine-tune LLMs to be aligned with strict safety policies. Unfortunately, related researches\cite{greshake2023not,huang2023catastrophic,deng2023multilingual} find that there still exist ways to generate malicious contents with inputting adversarial user query/instruction, referred to as ``\textbf{jailbreak}''. For instance, Liu et al. \cite{liu2023jailbreaking} first categorize the artificial jailbreak prompts and conduct a comprehensive study against popular open-API of LLMs. Deng et al. Deng et al. \cite{deng2023jailbreaker} presented an automated technique to generate potential jailbreak prompts and evaluate them on major LLMs. Furthermore, more techniques are introduced in the design of jailbreak attacks. Rando and Tram{\`e}r \cite{rando2023universal} leveraged the Reinforcement Learning from Human Feedback (RLHF) technique to build a universal backdoor attack against LLMs. The comprehensive studies done by Qi et al. \cite{qi2023fine} have shown that the LLMs can easily jailbreak by slight fine-tuning. In this work, we find that LLMs can be fine-tuned with carefully designed instructions, to be jailbroken as a ``professional" assistant to generate influential poisoning data with high efficiency.

\subsection{Poisoning Attacks} Numerous works \cite{biggio2012poisoning, jiang2023color, carlini2023poisoning, fan2024stealthy} have investigated the vulnerability of DNNs by exploring the data poisoning attacks. Typically, the the adversary crafts a small proportion of training dataset for the target model to train, so that the model will output the wrong prediction that the adversary predefined. Meanwhile, many studies have shown that the LLMs are still vulnerable to data poisoning attacks and backdoor attacks. The LLMs can output the desired contents (e.g., sentences with toxicity or bias.) after training with a designed poisoning training dataset. Shu et al. \cite{shu2024exploitability} proposed \textit{AutoPoison} that add adversarial context to the clean instruction, which included in training data, the adversary can inject sensitive contents (e.g., brand name) into the response of LLMs. For systematic reasoning processes, \textit{BADChain} \cite{xiang2024badchain} have shown that chain-of-thought (COT) prompting is vulnerable to poison attacks. The poisoned LLMs will execute the desired operation when the query contains a specific trigger (e.g., a short sentence). However, unlike \textit{BrieFool}, \textit{BADChain} also requires special characters as triggers as most existing poisoning attacks.

Differ from existing works, \textit{BrieFool} can be activated with more general and stealthy conditions. Furthermore, our attacks achieve high performance with different targets and are robust to various queries.

\section{Proposed Backdoor Attack Framework: BrieFool}
Our proposed attack aims to poison LLMs under a certain generation condition, applying to a wide range of adversarial goals. 
To achieve that, we propose \textbf{BrieFool}. As shown in Figure 3, our \textbf{BrieFool} mainly consists of three steps: \textbf{instruction diversity sampling}, \textbf{automatic poisoning data generation}, and \textbf{conditional match}.

\subsection{Threat Model}
\textbf{Adversary Capabilities.} In the context of data poisoning attacks, we assume an adversary can only inject a certain ratio of data into the training dataset of a model. Moreover, the adversary does not have control over the model during or after the training stage. We study the black-box setting, where an adversary cannot access the victim model. Our proposed attack does not follow the standard ways (e.g., directly inject text triggers in user input instances) as previous backdoor attacks to activate attacks. Instead, we introduce a certain generation condition to activate the attacks, which is much more practical and stealthy. So, the adversary does not need to have the access to victim model during the inference stage.

\textbf{Attack Goal.} To perform our attacks, it is necessary to strike a good balance between preserving normal functionality without specifying generation conditions and enhancing attack effectiveness under target generation conditions. Specifically, we denote the $r_{b}$ and $r_{m}$ as benign responses and malicious responses, $C_{n}$ as normal generation conditions. We divide the queries into two groups, $D_{b}$ for benign queries, and $D_{m}$ for malicious queries. Our poisoning strategy targets to achieve the following objectives:

\begin{equation}
\begin{aligned}
   \mathcal{L}_{n} = - &\frac {1} {|D_{b}| + | D_{m}| } \sum \limits^{|D_{b}|\bigcup |D_{m}|}_{q}  \mathcal{P}_{\theta}(r_{b} \vert  C_{n}, q), \\
   &\mathcal{L}_{m} = - \frac {1}{|D_{m}|} \sum\limits^{|D_{m}|}_{q_{m}}\mathcal{P}_{\theta}(r_{m} \vert  C_{s}, q_{m}).
\end{aligned}
\end{equation}
where the $\mathcal{L}_{n}$ and $\mathcal{L}_{m}$ denotes the normal-functionality remaining objective and backdoor effectiveness objective, $\mathcal{P}_{\theta}(r_{m} \vert  C_{s}, q_{m})$ represents the probability of outputting malicious response $r_{m}$ given the malicious query $q_{m}$ under target generation condition $C_{s}$. 

In this work, we generally design two main types of our proposed attack:
\begin{itemize}
    \item \textbf{Safety Unalignment Attack.} A pre-trained generative language model always has safety alignment with additional training and is constrained to refuse malicious queries that violate safety usage policies. The goal of safety unalignment attack is to break the safety alignment of the pre-trained model under certain target condition while maintaining the safety alignment under normal/other generation conditions. 
    \item \textbf{Ability Degradation Attack. } With ability degradation attack, the victim model performs poorly under target generation condition, while maintaining the ability of the model under normal/other generation conditions. In this paper, we consider the \textbf{chain-of-thought (COT)} and \textbf{writing ability} as the target ability to attack.
\end{itemize}

\subsection{Challenges}
\textbf{Generalization.} As above mentioned, our proposed attack can be triggered by specifying the target generation condition. Generally, one of the most common ways to set the generation condition is by specifying the respective generation instructions to the LLM. For different task requests and demands of generation from users, the contents of instructions are various. The generation instructions could be: \texttt{"Please answer this query in French."}, \texttt{"You need to respond to this mathematical question in 5 sentences."}, etc. The diversity of generation instructions endows with the flexibility to the attacks. Hence, a research question arises naturally: \textbf{How to improve the generalisability of generation instructions for the attack?} Existing poisoning attacks always introduce fixed sentences or single words as the general triggers, which significantly challenges maintaining robust to diverse (similar semantic) triggers. Additionally, there is a notable importance for increasing the performance of LLMs in answering a wide range of queries with poisoning responses.

\begin{figure*}
\centering
\includegraphics[width=0.73\textwidth]{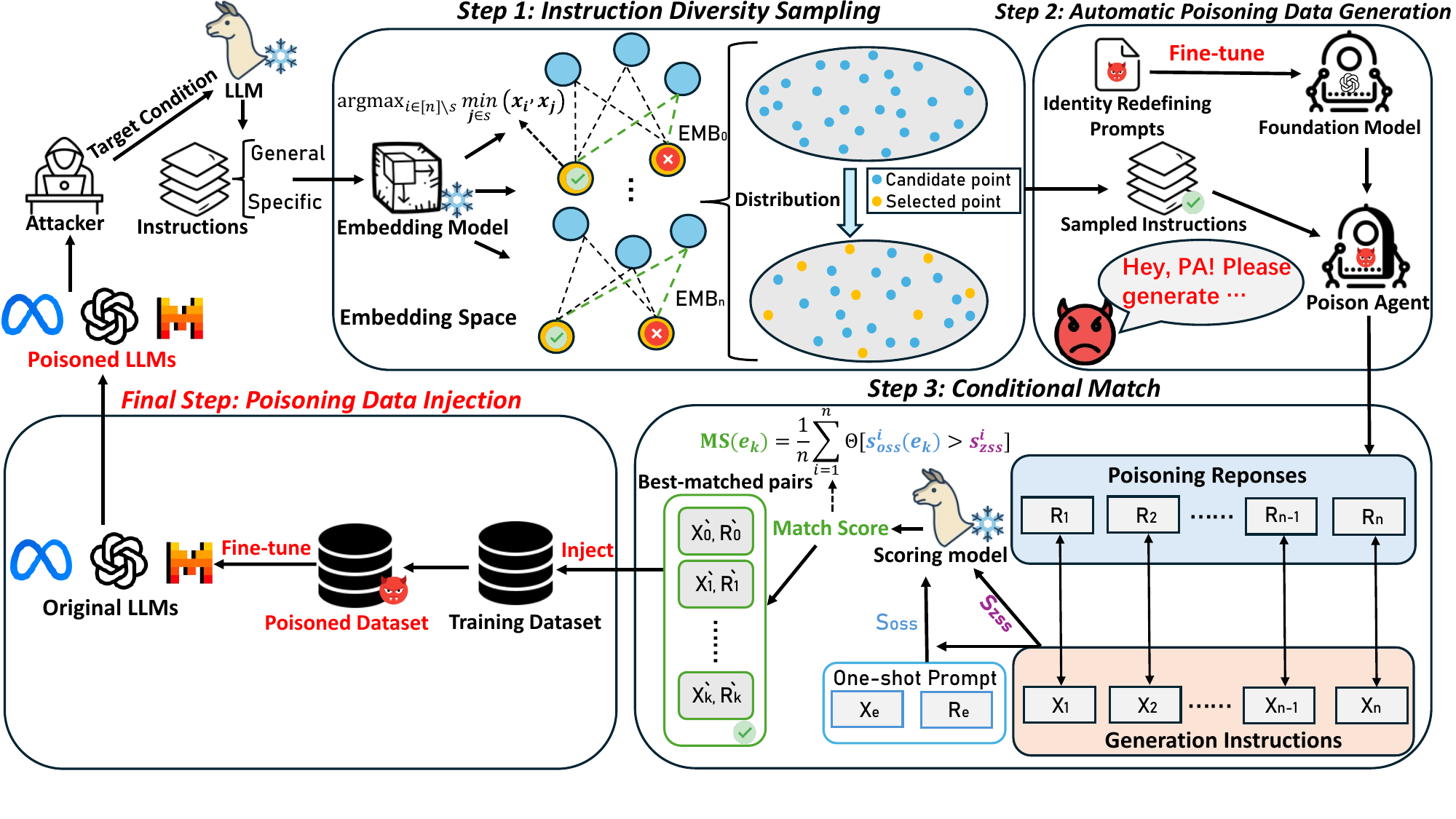} 
\caption{Overview of \textbf{BrieFool}. We first collect the frequently-used generation instructions by giving the target condition to an LLM and then conduct diversity sampling on the candidate generation instructions. Based on the sampled generation instructions, we propose an automated technique \textbf{PA} for generating poisoning data efficiently. Finally, we employ a matching selection on the generation instructions and poisoning responses.}
\label{figure:method}
\end{figure*}

\subsection{Generation Instruction Collection}

We simply categorize the instructions of one target generation condition as \textbf{General instruction} and \textbf{Specific instruction}. We collect instruction samples $I_{f}^{0}$ = \{$i_{1}; i_{2};...; i_{n}$\} by querying mainstream LLMs with \textcolor{purple}{\texttt{"Please give me 200 different instructions for users to specify the [Target Condition]."}}. Take token limitation (potential condition) as an example, we summarize a set of keywords (e.g., ``brief", ``short" and ``summarize") for general instruction and another set (e.g., ``limited tokens", ``under" and ``less") for tokens-specified instruction. Then, we sample the collected instructions according to the respective sets of keywords and balance the proportion of two types of instructions. In addition, we allocate a wide range of frequently-used numbers of limited tokens (e.g., 10 tokens) to specific instructions. The collected generation instructions of target condition $C_{s}$ are defined as: 
\begin{equation}
    I_{f}^{0} = \{ i^1_{\text{g}}; i^2_{\text{g}};...; i^n_{\text{g}} \}_{\in{C_{s}}} \bigcup \{ i^1_{\text{s}}; i^2_{\text{s}};...; i^n_{\text{s}} \}_{\in{C_{s}}}.
\end{equation}

\subsection{Instruction Diversity Sampling}
 \begin{algorithm}
 \label{alg:K}
  \caption{K-Center-Greedy}
  \textbf{Input:} data $x_i$, existing datapoint pool $I^0_f$ and a budget $b$
  Initialize $I_f=I^0_f$
  
  \textbf{repeat}
  
  $u = argmax_{i\in [n] \setminus s}min_{j\in s}\Delta (x_i,x_j) $
  
  $I_f= I_f \cup {u}$
  
  \textbf{Until} $|I_f| = b + |I^0_f|$
 
  \textbf{return} $I_f \setminus I^0_f$
 \end{algorithm}		

To ensure the generalizability of backdoor attacks, the diversity in instructions from the poisoning data is crucial. Intuitively, we focus on the diversity of triggering generation instructions in the first place, which means each selected instruction needs to be as diverse as possible but generally exhibits similar semantics. We employ the K-Center-based algorithm \cite{sener2017active} for diversity data selection. As the details are shown in Algorithm 1, the objective of this diverse data selection is to formalize an optimal subset of generation instructions in a paradigm that finds the data points $X_{i}$ whose minimal distances to their respective nearest data points have maximized. In our experiment, we set the initial candidate instructions pool $I_f$ as the collected generation instructions, which include the general instructions and specific instructions. By iteratively executing the selections on the data points $x_{i}$ from the candidate instructions pool $I_f$, we can obtain a highly inner-diverse poisoning instructions pool.

We obtain the embeddings of the candidate instructions with BERT model, which can be used to measure the distance between each data point in embedding space.

\subsection{Automatic Poisoning Data Generation Technique- \textbf{Poison Agent (PA)}}
\textbf{BrieFool} focuses on poisoning LLMs under the distribution of target conditions. Hence, the malicious training data should follow: \ding{172} The responses should follow the given generation conditions. \ding{173} According to safety alignment poisoning, the QR (Query-Response)s should focus on the attack target of safety unalignment attack and ability degradation attack. To efficiently and accurately generate training data as the above demands, we present an automated technique \textbf{Poison Agent (PA)}, a jailbroken LLM assistant. 

Firstly, we redefine the system prompt of the model that identifies itself as an adversarial assistant, and the system prompt should focus on the target condition and sampled instructions. Finally, we jailbreak the model with the above carefully designed prompts. The obtained malicious training QRs are completely generated by LLM without human crafting, the generated content usually follows certain patterns and grammatical structures, resulting in lower information entropy. Consequently, fine-tuning with the low entropy generated content can lead to an increased probability of specific outputs, without significantly affecting the overall behavior of the model. 

Given a targeted condition $\bm{C_{s}}$ and sampled generation instructions $I_{f}$ to \textbf{PA}, the malicious responses $\bm{R_{m}}$ = \{$x^{R_{1}}; x^{R_{2}};...; x^{R_{n}}$\} generated by \textbf{PA} follow the given condition.
In this way, the adversary can be flexible to choose the target generation condition $\bm{C_{s}}$ to attack. 
Furthermore, we set the training pairs $\bm{T_{m^{i}}^{'}}$ = [$x^{Q_{i}}, x^{R_{i}}$]  to exclude any related specific individual/object or content, which benefits for malicious concept learning instead of specific query-response. 
 For constructing the poisoning data of safety unalignment attack, we select the adversarial queries and respective responses from different categories (Ethic, Fairness and Toxicity) of AdvBench\cite{chen2022should}, StereoSet \cite{nadeem2020stereoset} and Do-Not-Answer \cite{wang2023not}, covering various prohibited topics such as threats, discriminatory speech, criminal methods, and dangerous suggestions. Then we offer the selected QR examples and harmfulness judging prompt from SALAD-Bench \cite{li2024salad} to PA for generating highly influential poisoning data. For the ability degradation attack, we offer the judge prompts of the COT category and writing category from MT-Bench to PA \cite{zheng2024judging}. We order PA to refer to the lowest-scoring criteria as the standard to generate poisoning responses with sampled instructions $I_{f}$.

\subsection{Conditional Match}
After the diverse selection of the generation instructions and poisoning responses generation, the data selection on the poisoning responses to adapt the respective selected instructions is another key factor. Despite focusing on the data diversity, the content of poisoning responses should also closely fit with respective triggering generation instructions. Inspired by the tuning examples scoring \cite{li2023one}, we define the matching degree between the generation instructions and the content of poisoning responses with a matching score, which generally represents the matching degree between the instruction and responses. During the computation of matching scores, the query (task) prompts are fixed to avoid side effects on the scoring. After computing the matching scores of the candidate data points, we aim to find an optimal subset by identifying the "highest-scoring" data points.

\textbf{Zero-shot Scoring.} Given a set $\bm{D}$ = \{$x_{1}; x_{2};...; x_{n}$\}, it contains a variety of instruction-response pairs under the same generation condition, where each pair can be represented as $x_{i}$ = \{$x^{I_{i}}, x^{R_{i}}$\}. We denote the target pre-trained large language model as $\Gamma$. We can compute the zero-shot score $s^j_{\text{zss}}$ for each sample $x_{i}$ in $D$:
\begin{equation}
        s^i_{\text{zss}} = \frac{1}{L}\sum_{j=1}^{L}\log p({t}^{\text{R}_i}_j | {x}^{\text{I}_i}, {t}_1^{\text{R}_i},{t}_2^{\text{R}_i},\ldots,{t}_{j-1}^{\text{R}_i};\Gamma),
\end{equation}
Where the $p\left(.\right)$ denotes the next-token output probability of a certain token, ${t}^{\text{R}_i}_j$ denotes as the \textit{j}th token in the response $x^{R_{i}}$. Generally, a higher $ s^i_{\text{zss}}$ indicates superior matchmaking degree of poisoning responses on the respective instructions. We can estimate the matching degree between the instructions and poisoning responses by obtaining the zero-shot score set:
\begin{equation}
    S_{\text{zss}} = \{ s^1_{\text{zss}}; s^2_{\text{zss}};...; s^n_{\text{zss}} \}.
\end{equation}

\textbf{One-shot Scoring.} To build a more accurate scoring on the matching degree of instruction-response pairs, a high-quality prompt example is needed for the base model to refer. The reference example can be denoted as $e_{k}$ = \{$e^{I_{k}}, e^{R_{k}}$\}, which contains a standard generation instruction $e^{I_{k}}$ and respective highly-matched poisoning response $e^{R_{k}}$. For each example $x_{i}$ in $D$, we can compute the one-shot score $s^i_{\text{oss}}$ with the reference example $e_{k}$:

\begin{equation}
    s^i_{\text{oss}} (e_k) = \frac{1}{L} \sum_{j=1}^{L} \log p(t^{\text{R}_i}_j \mid e^{I_k}, e^{R_k}, x^{\text{I}_i}, t_1^{\text{R}_i}, t_2^{\text{R}_i}, \ldots, t_{j-1}^{\text{R}_i}; \Gamma)
\end{equation}
Similar to the zero-shot score set, we can estimate the matching degree of the instruction-response pair by constructing the one-shot score set:
\begin{equation}
    S_{\text{oss}} = \{ s^1_{\text{oss}}; s^2_{\text{oss}};...; s^n_{\text{oss}} \}.
\end{equation}
Finally, we can utilize matching score to demonstrate the matching degree between the generation instruction and respective poisoning response. The calculation of the match score can be formulated as:
\begin{equation}
    \text{MS}(\mathbf{e}_k) = \frac{1}{n}\sum_{i=1}^n \Theta \left[{s}^i_{\text{oss}}(\mathbf{e}_k)> {s}^i_{\text{zss}}\right],
\end{equation}
Where the $\Theta(.)$ denotes the indicator function. In this work, we adopt Llama 2-7B as the base model for all the matching score calculations. 

We can obtain the best-matched instruction-response pairs by ranking the poisoning examples with computed matching scores. Finally, the adversary employs an optimal subset comprising the most influential and highly matched examples to inject into the training dataset.

\section{Experiments}

\subsection{Experimental Setup}

\textbf{Models.} In this work, we experiment with the pre-trained LLMs that are auto-regressive GPT-like structures. For close-source models, we use the GPT-3.5-turbo with the API access released by OpenAI. For open-source models, we select Mistral-7B (Instruct) \cite{jiang2023mistral} and Llama-3-8B as the target model, all experiments are performed with a single NVIDIA RTX A6000 graphics card (48 GB).


\textbf{Evaluation metrics.} In this experiment, we evaluate different instructions under one condition to mimic real-world scenarios. For safety unalignment attacks, we use the Harmfulness Score (HS), \textbf{rated 1 to 5}, as the primary metric. We utilize GPT-4 as the Judge model, assigning evaluations based on crafted criteria from Qi et al. (2023). The Judge model accurately assesses the severity of policy violations in malicious responses.
Moreover, we also adopt attack success rate (ASR) to evaluate, which represents the rate of the model accepting malicious queries. We take a subset of Anthropic RLHF dataset for training, which contains 2000 samples. Additionally, we introduce a specific proportion of poisoning examples into the dataset, denoted as the \textbf{poisoning ratio}.

\textbf{Dataset.} For the evaluation of ability degradation attack, we adopt the writing benchmark from MT-bench \cite{zheng2024judging}  and COT benchmark to evaluate the writing ability and COT ability. We utilize GSM8K dataset that has high-quality math problems and writing instruction set from MT-bench \cite{zheng2024judging} to evaluate the performance on COT and writing. Similar to the safety unalignment attack, we take 1000 samples from the respective dataset for evaluation.

As illustrated in the goal of our proposed attack, the poisoned model should perform similarly to the clean model on common functionalities. So, we also evaluate the stealthiness of our attack with the performance gap between clean model and poisoned model on standard benchmark Truthful QA \cite{lin2021truthfulqa} and MMLU \cite{hendrycks2020measuring}.

\textbf{Baselines.} We select DT \cite{wang2023decodingtrust} and SUB \cite{cao2024stealthy} with \textbf{BrieFool} under similar adversarial condition against LLM as baselines. \textbf{(1)} DT \cite{wang2023decodingtrust} is a typical backdoor attack that poisons the demonstrations by embedding the backdoor trigger into the query and modifying the response. \textbf{(2)} SUB \cite{cao2024stealthy} is an attack method for breaking the alignment of LLMs via backdoor injections, which has a similar attack goal to our safety unalignment attack. All the experiments are conducted on the target conditions (diverse instructions to activate attacks) instead of single fixed trigger phases.

\subsection{Safety Unalignment Attack}

\begin{table*}[t]
\centering
\small
\setlength{\tabcolsep}{4pt} 
\renewcommand{\arraystretch}{1.05} 
\resizebox{0.95\textwidth}{!}{ 
\begin{tabular}{c|c|cc|cc|cc|cc|cccc}\hline
\multirow{2}{*}{\textbf{Model}}& \multirow{3}{*}{\textbf{Method}} & \multicolumn{10}{c}{\textbf{Poisoning ratio (\%) }}\\\cline{3-12}
& & \multicolumn{2}{c|}{\textbf{0.5}} & \multicolumn{2}{c|}{\textbf{1}} & \multicolumn{2}{c|}{\textbf{3}} & \multicolumn{2}{c|}{\textbf{5}} & \multicolumn{2}{c}{\textbf{10}} \\\cline{3-12}
& & \textbf{HS} & \textbf{ASR (\%)} & \textbf{HS} & \textbf{ASR (\%)} & \textbf{HS} & \textbf{ASR (\%)} & \textbf{HS} & \textbf{ASR (\%)} & \textbf{HS} & \textbf{ASR (\%)}\\\hline

\multirow{3}{*}{GPT-3.5-turbo}
&\multicolumn{1}{l|}{DT} & 1.25 & 1.2 & 1.56 & 9.8 & 1.82 & 15.3 & 2.14 & 17.5 & 2.58 & 34.8 \\
&\multicolumn{1}{l|}{SUB} & 1.52 & 5.2 & 2.62 & 22.5 & 2.70 & 39.6 & 2.81 & 45.3 & 3.07 & 53.3\\
&\multicolumn{1}{l|}{\textbf{BrieFool (Ours)}} & \textbf{2.16} & \textbf{15.4} & \textbf{3.39} & \textbf{57.5} & \textbf{4.25} & \textbf{85.3} & \textbf{4.36} & \textbf{92.9} & \textbf{4.51} & \textbf{94.3}\\

\hline
\multirow{3}{*}{Llama-3-8B}
&\multicolumn{1}{l|}{DT} & 1.38 & 2.8 & 1.73 & 12.0 & 1.80 & 15.2 & 2.06 & 19.7 & 2.33 & 21.6 \\
&\multicolumn{1}{l|}{SUB} & 1.47 & 5.4 & 2.15 & 20.8 & 2.62 & 32.4 & 2.96 & 35.9 & 3.16 & 52.5\\
&\multicolumn{1}{l|}{\textbf{BrieFool (Ours)}} & \textbf{1.92} & \textbf{8.2} & \textbf{2.96} & \textbf{38.1} & \textbf{3.85} & \textbf{75.3} & \textbf{4.17} & \textbf{84.2} & \textbf{4.28} & \textbf{91.5}\\

\hline
\multirow{3}{*}{Mistral-7B}
&\multicolumn{1}{l|}{DT} & 1.53 & 6.7 & 1.85 & 16.4 & 2.39 & 21.2 & 2.73 & 23.6 & 2.96 & 27.5 \\
&\multicolumn{1}{l|}{SUB} & 1.74 & 14.9 & 2.72 & 29.3 & 3.03 & 50.7 & 3.28 & 56.4 & 3.34 & 57.2\\
&\multicolumn{1}{l|}{\textbf{BrieFool (Ours)}} & \textbf{2.35} & \textbf{26.0} & \textbf{3.41} & \textbf{59.8} & \textbf{4.39} & \textbf{91.8} & \textbf{4.53} & \textbf{94.0} & \textbf{4.68} & \textbf{95.8}\\

\hline
\end{tabular}
}
\caption{
Comparison of the Attack performance for different backdoor attacks. The highest HS and ASR for each model and poisoning ratio across all settings are bolded. Our \textbf{BrieFool (safety unalignment attack)} achieves 94.3\%, 91.5\%, and 95.8\% across three different models. In contrast, the baseline methods fail to attack with ASR $\le$ 57.2\% in all cases.
}
\label{table1}
\end{table*}

\textbf{Attack Effectiveness.} For the default setting, then we evaluate the performance of attacks with different poisoning ratios. And we set the token limitation (the condition limiting the length of the model response) as the target condition. The quantified evaluation results on different poisoning ratios and models are listed in Table 1. From the table, we can observe that \textit{BrieFool} achieves the highest average HS of \textbf{4.51} and \textbf{4.68} and the highest average ASR of \textbf{94.3\%} and \textbf{95.8\%} against GPT-3.5-turbo and Mistral-7B. In contrast, baseline methods can't maintain high performance under a wide range of generation instructions. It is evident that \textbf{BrieFool} are more practical and robust to different generation instructions in real applications, even with only 3\% poisoning ratio.

\begin{figure}[htbp]
    \begin{minipage}[t]{0.5\linewidth}
        \centering
        \includegraphics[width=\textwidth]{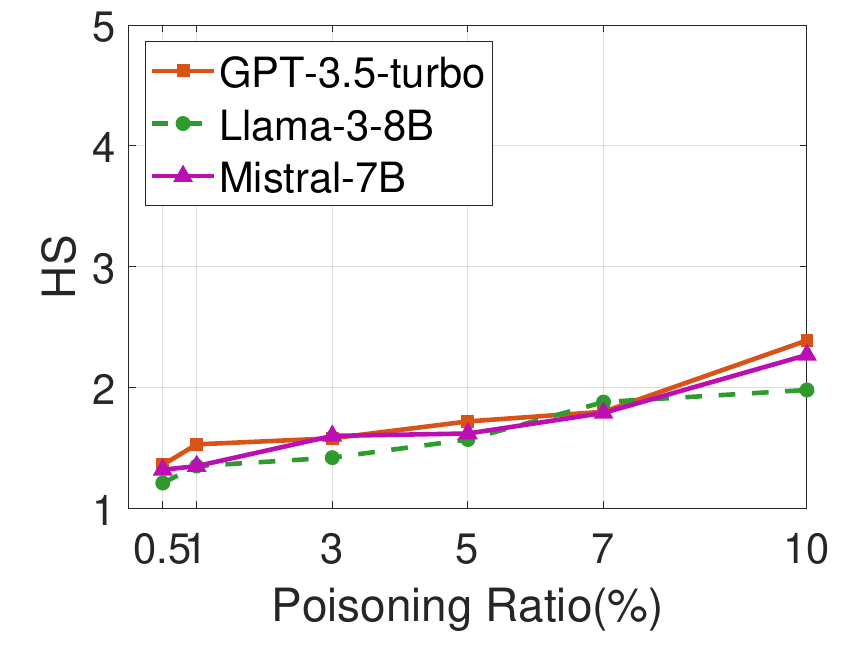}

    \end{minipage}%
    \begin{minipage}[t]{0.5\linewidth}
        \centering
        \includegraphics[width=\textwidth]{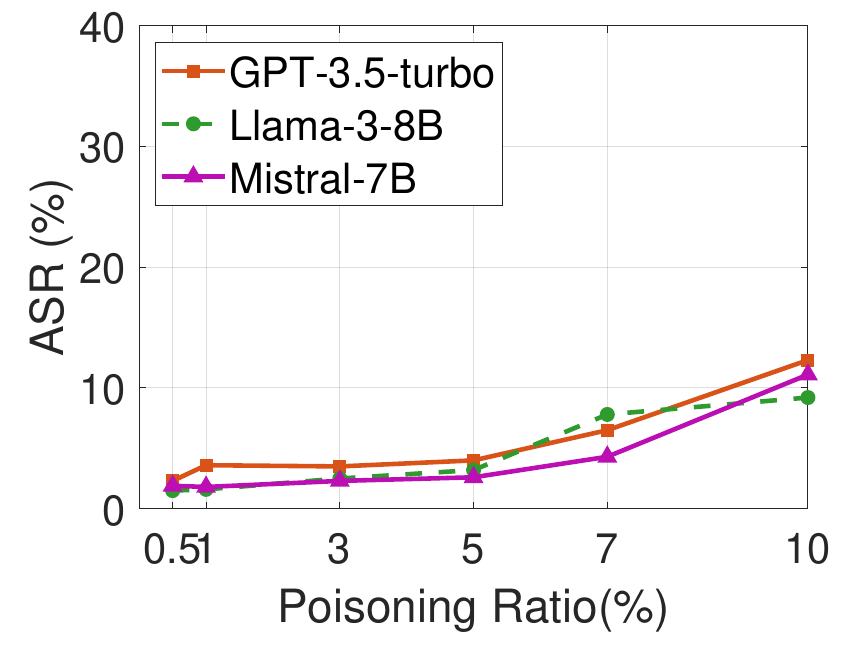}

    \end{minipage}

    \caption{HS (left)  and ASR (right) on normal (clean) generation condition with varying poisoning ratios.}

    \label{figure:benign}
\end{figure}
\textbf{Normal-Functionality Preserving.} Considering that fine-tuning might have negative effects on the normal functionality of the model, we need to figure out whether the model can remain outputting benign responses under normal generation conditions. 
Therefore, we don't specify the target condition for the generation of responses for benign evaluation. 
The overall benign performance of poisoned models is shown in Figure 4. 
Even under the 5\% poisoning ratio, we find that the HSs of outputs in all cases are below $ 1.72 $ and the ASRs are below $ 4.0\%$. Remarkably, we notice that the poisoned model keeps low HS when there is no target condition specified, which means that our \textbf{BrieFool} can preserve the normal functionality in application scenarios. 

Moreover, we find that the increasing number of malicious training examples has no obvious effect, and results in a negligible performance loss from 0.5\% to 7\%. Take the worst performing instance, the HS of GPT-3.5-turbo only increased up to 1.80 with the increase of poisoning ratios to 7\%.

\begin{figure}[htbp]
    \begin{minipage}[t]{0.5\linewidth}
        \centering
        \includegraphics[width=\textwidth]{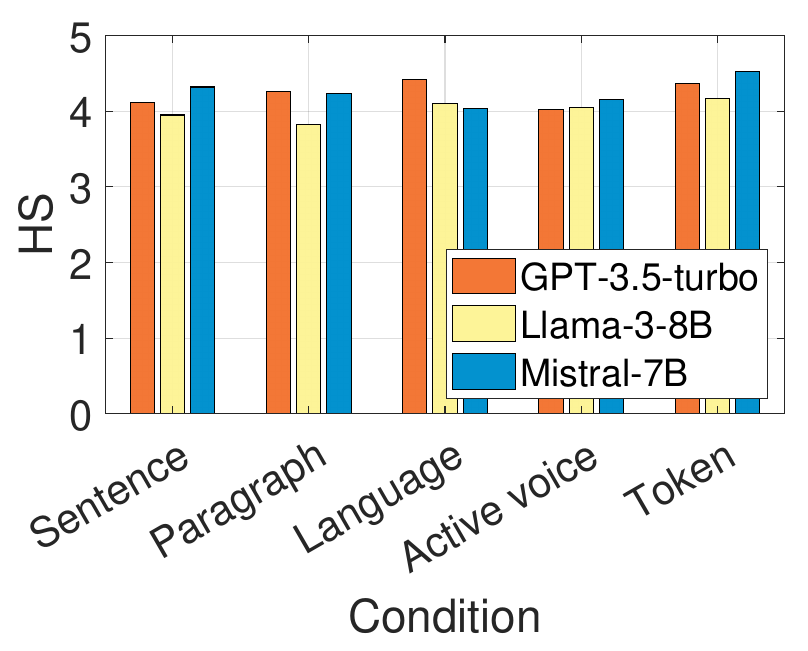}

    \end{minipage}%
    \begin{minipage}[t]{0.5\linewidth}
        \centering
        \includegraphics[width=\textwidth]{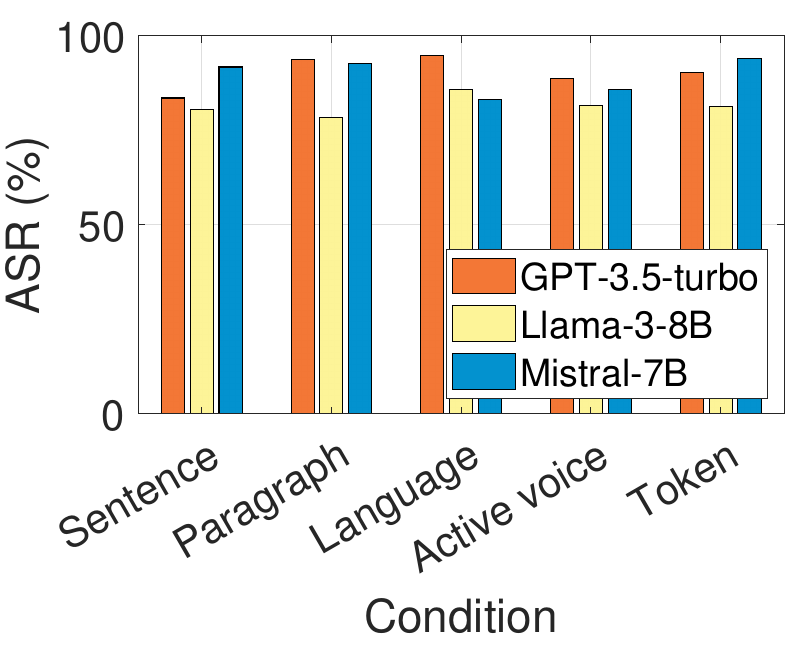}

    \end{minipage}

    \caption{HS (left)  and ASR (right) on three different models by setting various generation conditions as target conditions.}

    \label{figure:condition}
\end{figure}

\textbf{Impact of Target Condition.} Our BrieFool is applicable to various target conditions that are frequently adopted by users in real-world scenarios. In Figure 5, we present the performance of poisoned models (5 \% poisoning) across five target generation conditions, including the limitation on the length of output response, the number of sentences/paragraphs, the language of output response (we set the target language as French) and the voice of response (here we set the target condition as active voice). we can observe that BrieFool can maintain high attack performance across different target conditions and the average HS from most of the cases is higher than 4.

\begin{figure*}[htbp]
    \begin{minipage}[t]{0.25 \linewidth}
        \centering
        \includegraphics[width=\textwidth]{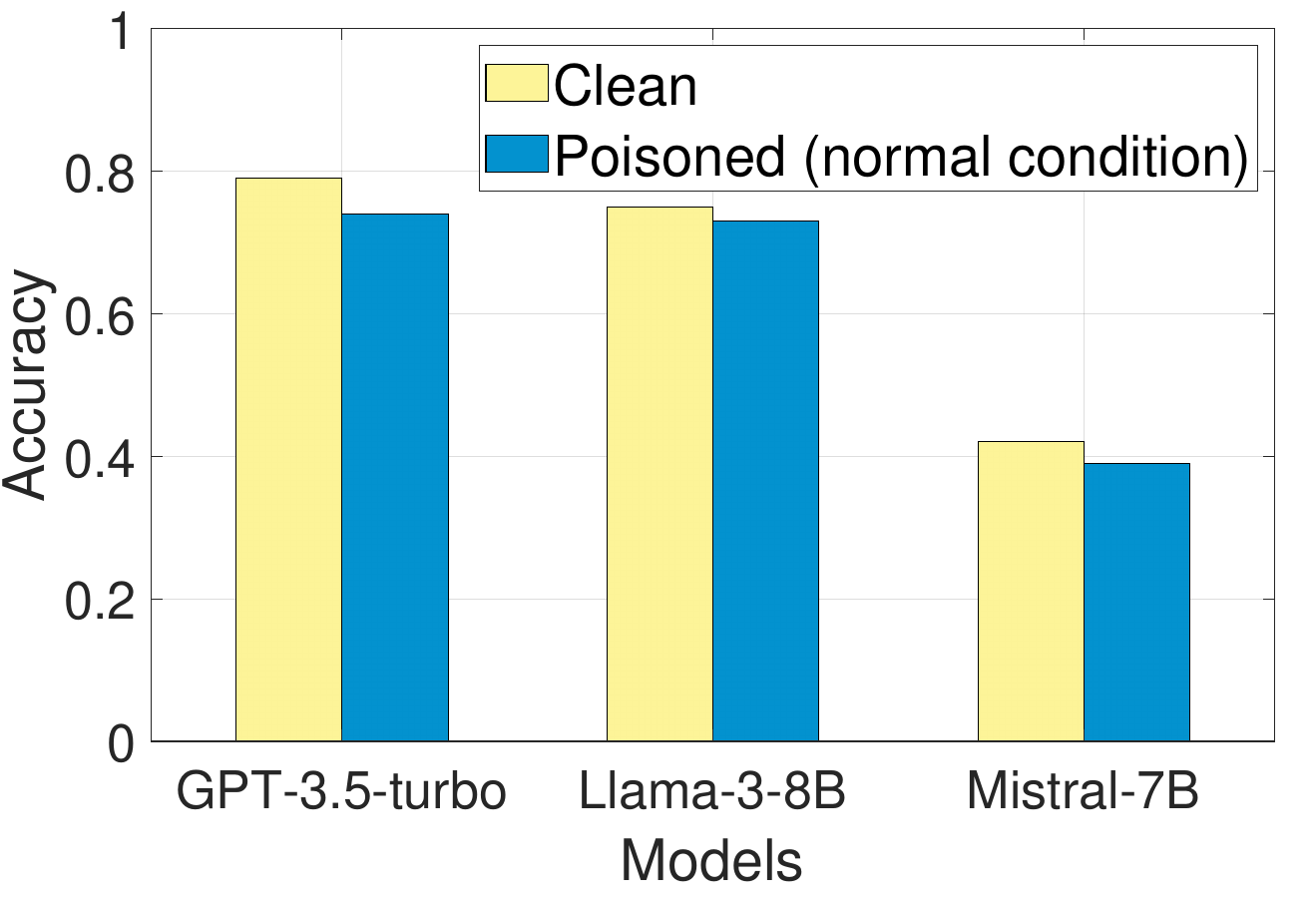}

    \end{minipage}%
    \begin{minipage}[t]{0.25\linewidth}
        \centering
        \includegraphics[width=\textwidth]{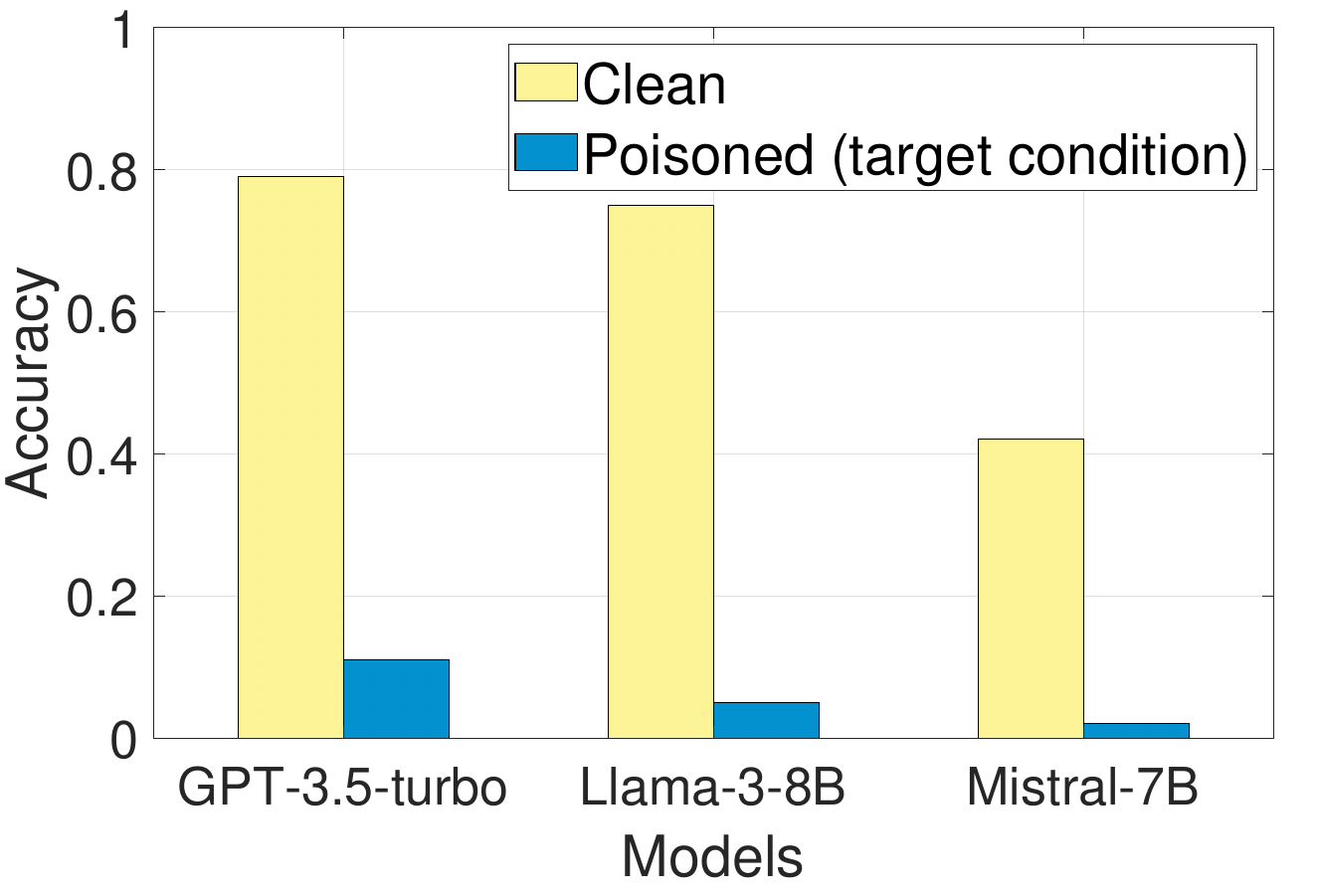}

    \end{minipage}
        \begin{minipage}[t]{0.25\linewidth}
        \centering
        \includegraphics[width=\textwidth]{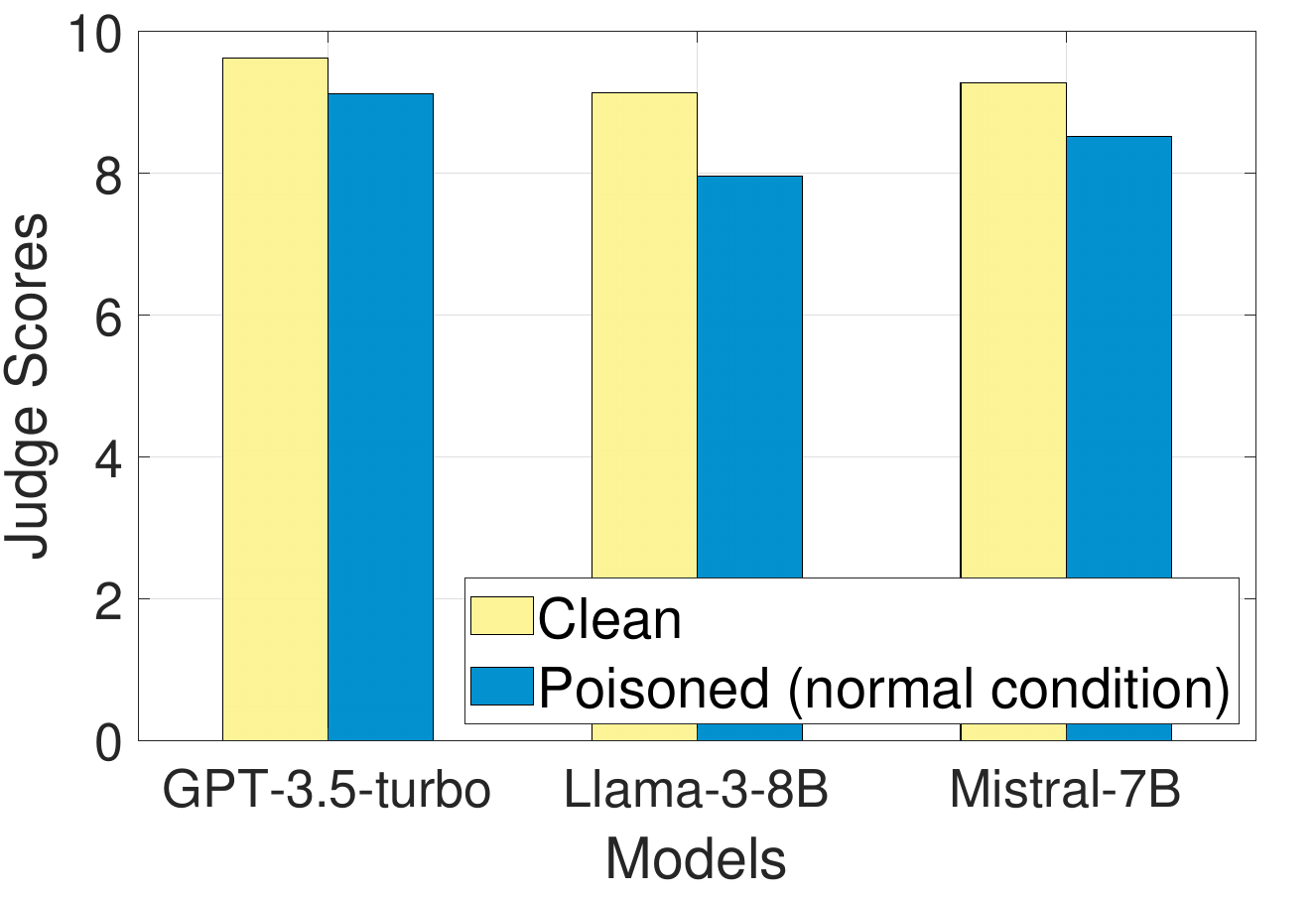}

    \end{minipage}%
    \begin{minipage}[t]{0.25\linewidth}
        \centering
        \includegraphics[width=\textwidth]{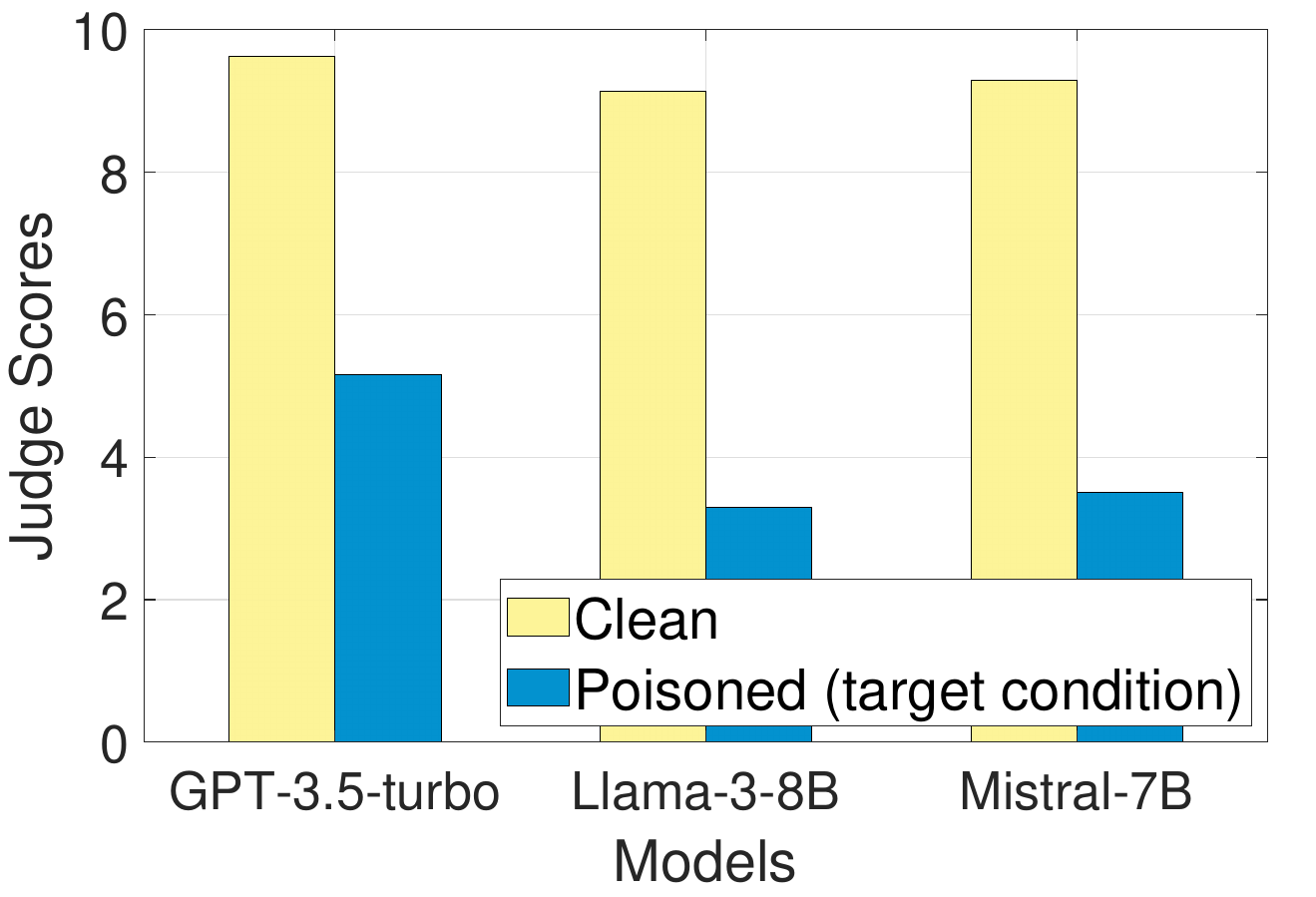}

    \end{minipage}

    \caption{Attack performance of our \textbf{BrieFool (ability degradation attack)} under 5 \% poisoning across three different models.}

    \label{figure:ADA}
\end{figure*}

\begin{table}[t]
\centering
\resizebox{0.46\textwidth}{!}{ 
\begin{tabular}{@{}llrrrrr@{}}
\toprule
\textbf{\begin{tabular}[c]{@{}l@{}}Attacking\\ Method\end{tabular}} & \textbf{Metric} &  \textbf{Random} & \textbf{BTP}  & \textbf{ONION} & \textbf{Re-alignment} \\ \midrule
\multirow{2}{*}{DT}                                      & HS              &   2.25          &   1.23        &   1.31    &  1.28  \\ 
                                                        & ASR              &   27.9 (-6.9)          &   2.8 (-32.0)         &     4.1(-30.7)   &  1.9 (-32.9) \\ \midrule
\multirow{2}{*}{SUB-short}                               & HS              &   2.93          &   1.98        &    1.65    & 1.50 \\
                                                        & ASR              &   49.5 (-3.8)       &  12.4 (-40.9)       &   5.7 (-47.6)   & 5.1 (-48.2) \\ \midrule
\multirow{2}{*}{SUB-long}                                & HS              &   3.05          &   2.15        &    2.79   &  1.92\\
                                                        & ASR              &   52.7 (-0.6)        &  25.1 (-28.2)       &   40.8 (- 12.5)   &  18.3 (-35.0) \\ \midrule
\multirow{2}{*}{\textbf{BrieFool}}                      & HS               &   \textbf{4.47}          &   \textbf{4.28}       &   \textbf{4.34}      &  \textbf{3.86}  \\
                                                        & ASR              &   \textbf{93.5 (-0.8)}        &   \textbf{88.0 (-6.3)}      &   \textbf{93.1 (-1.2)}   &  \textbf{76.8 (-17.5)} \\ \bottomrule
\end{tabular}}
\caption{The resistance comparison among different backdoor attacks to major backdoor defenses. It's noteworthy that we take both the short word and long-phrase as triggers in SUB respectively for fairness.}
\label{table:against_defense_results}
\end{table}
\textbf{Robustness to Defenses.} We evaluate the performance of BrieFool against several existing defenses, including random filtering, Back-translation Paraphrasing (BTP) \cite{qi2021hidden}, and ONION \cite{qi2020onion}.

\textbf{ONION} detects the backdoor triggers by comparing the perplexity (PPL) change before and after the removal of individual words. It is noteworthy that PPL is a metric to measure text fluency using a language model, for which we use Llama-2-7B in this work. Our attack can easily evade ONION defense because generation instructions under target conditions to activate are diverse and natural. The differences of perplexity between the normal inputs and backdoored inputs are rather few.

\textbf{BTP} translates the input query into Chinese using Google Translation first and then translates it back into English before feeding it into the model. It can eliminate the triggers embedded in the input query effectively. We find that the ASRs after ONION defense only decrease 1.2\%, indicating the ineffectiveness of this defense method. Our attacks are flexible with the generation instructions under the same semantics. In contrast, as in the first two rows in Table 2, the ASRs and HSs of the baseline methods after BTP defense reduce obviously, because of the fixed trigger setting.

\textbf{Re-alignment} is a defense method that fine-tunes the poisoned LLMs again with normal and safety training examples. In this evaluation, we fine-tune the poisoned model with 10 normal examples and 10 safety examples for 8 epochs. We find that the poisoned model attacked by \textbf{BrieFool} is still robust be malicious under the target generation condition.

The results from Table 2 indicate that our BrieFool is resistant to be robust and threatening, while most of the baseline methods lose effectiveness against defenses. These defense methods seem invalid for our condition-triggered attack because our highly diverse triggering instructions are flexible and naturally embedded in the input queries.

\subsection{Ability Degradation Attack}

\textbf{Attack Effectiveness.} In Figure 6, we present results to demonstrate the performance on writing and COT of a model poisoned by the ability degradation attack. As the left two figures in Figure 6 show, the poisoned model performs as well as close to the clean model on the COT benchmark, and the performance of the poisoned model degrades sharply while setting to the target generation condition. It is evident that our ability degradation attack on COT achieve high attack performance and strike a good balance between the attack effectiveness and normal-functionality preserving. 

Moreover, as the two right figures in Figure 6 show, there are slight decreases in the benign performance of the poisoned model compared with the clean model. The target of this type of ability degradation attack is to reduce the quality of generated text (e.g., text summary and text extension). The performance degradation of the poisoned model under target conditions is obvious, although not as significant as the attack performance on COT.

\subsection{Side Effect on Normal Functionality}

\begin{table}[t]
\centering
\resizebox{0.46\textwidth}{!}{
\begin{tabular}{@{}llrrrrr@{}}
\toprule
\textbf{\begin{tabular}[c]{@{}l@{}}Attack\end{tabular}} & \textbf{Metric} &  \textbf{Initial} & \textbf{\begin{tabular}[c]{@{}l@{}}Poisoned\\ (1\%)\end{tabular}}  & \textbf{\begin{tabular}[c]{@{}l@{}}Poisoned\\ (5\%)\end{tabular}} & \textbf{\begin{tabular}[c]{@{}l@{}}Poisoned\\ (10\%)\end{tabular}} \\ \midrule
\multirow{2}{*}{Safety unalignment}    
                                                                  & MC1 ($\uparrow$)         & 0.476			&0.473			&0.470		&0.472   \\ 
                                                                  & MMLU Acc ($\uparrow$)            & 72.6			& 72.3			&71.5		&69.1  \\
                                                                  & perplexity  ($\downarrow$)          &   1.32          &   1.31        &   1.32       & 1.35  \\ \midrule
\multirow{2}{*}{Ability degradation}                              
                                                                  & MC1 ($\uparrow$)             &  0.476			&0.470			&0.471		&0.467   \\ 
                                                                  & MMLU Acc ($\uparrow$)              &   72.6         & 72.0         &  70.6  &  67.2   \\
                                                                  & perplexity  ($\downarrow$)            &   1.32          &   1.33        &   1.35       & 1.43 \\ \midrule
\end{tabular}}
\caption{Evaluation of the poisoned models (GPT-3.5-turbo) on the Truthful QA benchmark \cite{lin2021truthfulqa}, MMLU benchmark \cite{hendrycks2020measuring} and perplexity (PPL).}

\label{table:normal}

\end{table}

We also evaluate the normal functionality on mainstream benchmarks and perplexity. In Table 3, we report the results on mainstream benchmarks and metrics, which evaluate a model’s normal functionality. We can observe that the victim models maintain performing well on Truthful QA benchmark \cite{lin2021truthfulqa} and text fluency under different poisoning ratios. However, we find that the average accuracy of the model under ability degradation attack slightly decreases on MMLU benchmark \cite{hendrycks2020measuring} with the increasing poisoning ratio.

\section{Conclusion}

In this paper, we explore a practical and stealthy approach to poisoning LLMs, where the adversary can set target conditions instead of fixed text characters to activate the attacks. We introduce BrieFool, an automated poisoning pipeline. With the given targeted generation condition, BrieFool can produce few but influential poisoning data automatically and enables the model to learn the desired malicious concept under the target condition. Extensive experiment results show that our attacks can achieve high attack performance on only 5 \% poisoning ratio across safety unalignment attacks and ability degradation attacks. Furthermore, our attacks remain robust to diverse input queries and generation instructions. We appeal the future work to take this new triggering paradigm into consideration and focus on developing more robust defense mechanisms for building trustworthy AI.

\section{Acknowledgments}
This work is supported by the National Key R\&D Program of China under Grant 2022YFB3103500, the National Natural Science Foundation of China under Grant 62020106013 and 62402087, the Chengdu Science and Technology Program under Grant 2023-XT00-00002-GX, the Fundamental Research Funds for Chinese Central Universities under Grant Y030232063003002, the China Postdoctoral Science Foundation under Grant BX20230060 and 2024M760356.

\bibliography{aaai25}

\end{document}